\documentclass[sigconf,nonacm]{acmart}

\usepackage{graphicx}
\usepackage{algorithm}
\usepackage{algpseudocode}
\usepackage{multirow}
\usepackage{pifont}

\usepackage{nicematrix}
\usepackage[table]{xcolor}
\usepackage{booktabs}
\usepackage{booktabs}
\usepackage{multirow}
\usepackage{graphicx}
\usepackage{hyperref}
\definecolor{headergray}{gray}{0.9}
\definecolor{rowblue}{rgb}{0.92, 0.95, 1}

\usepackage{xspace}
\newcommand{\ours}{\texttt{HTDC}\xspace}

\begin{document}

\title{HTDC: Hesitation-Triggered Differential
Calibration for Mitigating Hallucination in Large Vision-Language Models}

\author{Xinyun Liu}
\email{2023141010186@stu.scu.edu.cn}
\affiliation{%
  \institution{Sichuan University}
  \city{Chengdu}
  \country{China}}

\renewcommand{\shortauthors}{Xinyun Liu}

\begin{abstract}
Large vision-language models (LVLMs) achieve strong multimodal performance, but still suffer from hallucinations caused by unstable visual grounding and over-reliance on language priors. Existing training-free decoding methods typically apply calibration at every decoding step, introducing unnecessary computation and potentially disrupting stable predictions. We address this problem by identifying layer-wise hesitation, a simple signal of grounding instability reflected by fluctuations in token preference across intermediate layers. Based on this observation, we propose Hesitation-Triggered Differential Calibration (\ours), a training-free decoding framework that preserves standard full-branch inference and activates calibration only at hesitation-prone steps. When triggered, \ours contrasts the full branch with two lightweight probes, a visual-nullification probe and a semantic-nullification probe, to suppress hallucination-prone candidates while avoiding unnecessary intervention on stable steps. Experiments on representative hallucination benchmarks show that \ours consistently reduces hallucinations while maintaining strong task accuracy, achieving a favorable trade-off between effectiveness and computational overhead.
\end{abstract}



\maketitle

\begin{figure}[t]
  \centering
  \includegraphics[width=\linewidth]{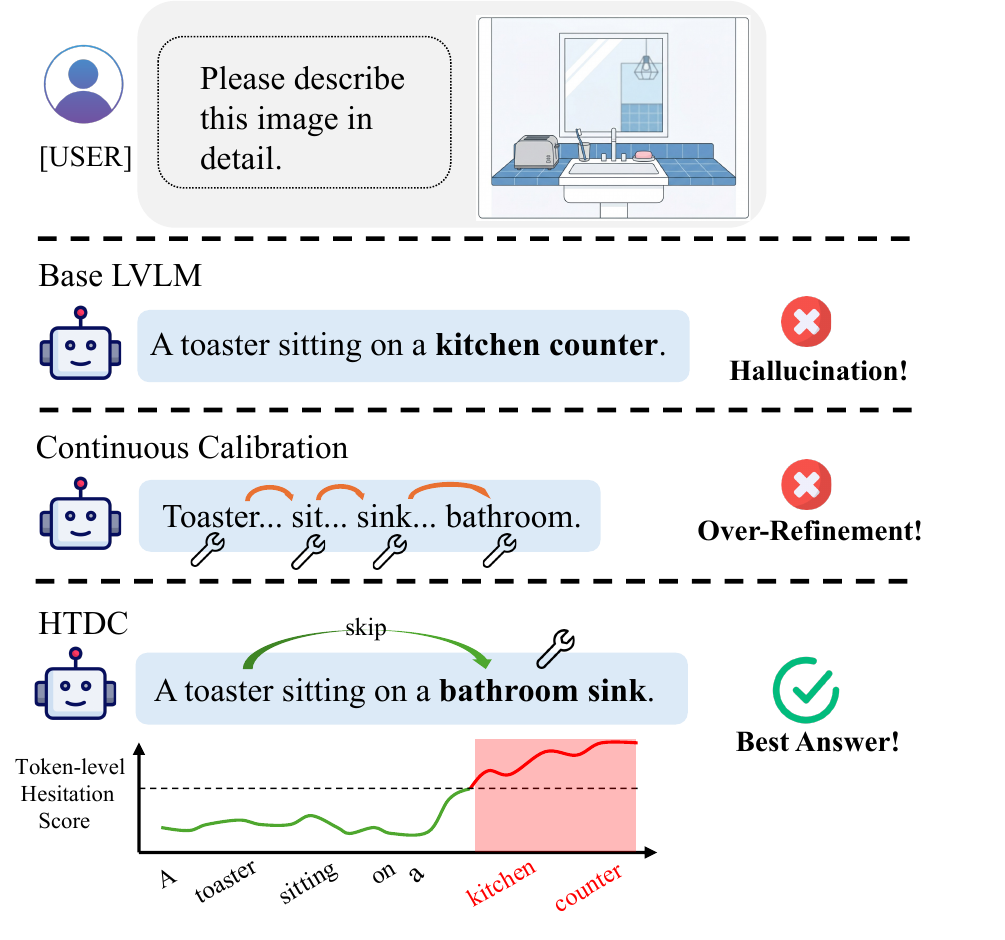}
   \caption{Motivation of our method. The base LVLM hallucinates the scene context and incorrectly describes the toaster as being on a kitchen counter. Continuous calibration refines all tokens indiscriminately, leading to over-refinement without correcting the error. In contrast, our \ours selectively calibrates only hallucination-prone tokens based on token-level hesitation scores, yielding a more accurate description.}
  \label{fig:motivation0}
\end{figure}

\section{Introduction}
Large Vision-Language Models (LVLMs) have shown impressive capabilities in image understanding and visual question answering~\cite{lee2018pre, liu2023visual, liu2024improved, wang2024qwen2, bai2025qwen3}. However, their responses have always been plagued by hallucinations, referring to the phenomenon where the model generates textual descriptions that are coherent but inconsistent with the provided visual input or deviate from objective facts~\cite{maynez2020faithfulness, huang2025survey, yin2024woodpecker, bai2024hallucination, liu2024survey}. This issue is particularly problematic in safety-critical or decision-making scenarios, where reliability and factual consistency are essential~\cite{yin2024survey, li2024manipllm, ma2024coco, hager2024evaluation, huang2025survey, kang2023deficiency}.

Existing work can be broadly categorized into two paradigms. Training-based approaches aim to rectify the model's internal representations by optimizing its parameters, which involves supervised fine-tuning (SFT) and reinforcement learning from human feedback (RLHF)~\cite{wang2024mitigating, yu2024rlhf,yang2025mitigating, zhao2024mitigating, zhou2024aligning}. In contrast, training-free decoding-time approaches have gained increasing attention due to their lower computational cost and stronger practicality, as can be directly applied to existing LVLMs during inference. A common strategy is contrastive decoding, which compares model outputs under different conditions (e.g., clean vs perturbed visual inputs) and then adjust token probabilities accordingly~\cite{leng2024mitigating,chuang2023dola, wang2024mitigating, lee2024delve, zhang2025active, park2025convis}. Another line of work exploits the model's internal dynamics, such as intermediate-layer signals, to identify unstable generation and refine predictions for hallucination mitigation~\cite{chuang2023dola, mir2025geometry, wang2025damo, yu2025hallurnn, zhang2025active}.  


Despite their effectiveness, existing decoding-time methods typically refine the prediction of every generated token. However, such uniform intervention may be unnecessary, since not all tokens contribute equally to visually grounded reasoning. For example, function words such as prepositions (e.g., ``at'') are usually less relevant to visual grounding~\cite{wang2025damo}, whereas hallucinations are more often associated with semantically important tokens, particularly those describing attributes such as color, position, and count~\cite{wang2025damo}. Applying refinement indiscriminately to all tokens therefore introduces unnecessary computational overhead. For instance, VCD~\cite{leng2024mitigating} requires two forward passes to obtain the logits for a single token, resulting in nearly doubled decoding cost. Moreover, uniform refinement may lead to over-refinement, which can even degrade generation quality in some cases. As illustrated in Figure~\ref{fig:motivation0}, we provide a representative example in which the base LVLM produces a hallucinated response, while the decoding-time calibration method, instantiated here with VCD, over-corrects the generation and still fails to resolve the hallucination effectively.

To address these limitations, we propose \textbf{H}esitation-\textbf{T}riggered \textbf{D}ifferential \textbf{C}alibration (\ours), a training-free decoding framework that preserves the standard full-branch decoding process and invokes calibration only when intervention is necessary. Concretely, \ours leverages layer-wise hesitation, an observable signal derived from internal layer dynamics, to identify decoding steps that are more likely to be hallucination-prone. Once such hesitation is detected, \ours performs hesitation-triggered differential calibration by contrasting the full branch with two probe branches: a visual-nullification branch and a semantic-nullification branch. This contrastive design explicitly exposes candidates whose scores remain spuriously high even when visual evidence or query semantics is removed, and suppresses them during decoding. In this way, \ours reduces unnecessary computation, minimizes interference with the model’s normal decoding behavior on stable steps, and provides a more interpretable mechanism for hallucination mitigation.

Extensive experiments on two LVLMs across four hallucination benchmarks show that \ours significantly mitigates hallucinations and achieves superior performance over existing challenging baselines. The contributions of this paper are summarized as follows:

\begin{itemize}
    \item We introduce layer-wise hesitation, a simple yet effective metric derived from internal decoding dynamics, to characterize grounding instability and identify hallucination-prone decoding steps in LVLMs.

    \item We propose Hesitation-Triggered Differential Calibration (\ours), a training-free decoding framework that preserves standard full-branch inference and activates calibration only at hesitation-prone decoding steps. \ours uses two lightweight probe branches, a visual perturbation probe and a semantic nullification probe, to perform differential calibration in an interpretable manner.

    \item Extensive experiments on representative hallucination benchmarks show that \ours consistently mitigates hallucinations while maintaining strong task accuracy, achieving a favorable trade-off between effectiveness and computational overhead.
\end{itemize}

\section{Related Work}
\subsection{LVLMs and Hallucination}
Large Vision-Language Models (LVLMs) typically consist of three components: a vision encoder, a projector that maps visual features into the language embedding space, and a Large Language Model (LLM) that performs autoregressive text generation. Representative systems such as LLaVA adopt this architecture to connect CLIP-ViT visual representations with an LLM through a lightweight projector module~\cite{liu2023visual,liu2024improved}, while more recent models such as Qwen3-VL further extend this paradigm with dynamic-resolution visual tokenization~\cite{wang2024qwen2, bai2025qwen3}. Given an image and a textual instruction, an LVLM first encodes the image into visual tokens, projects them into the language embedding space, and then generates the response token by token conditioned on both visual and textual context. Based on this architecture, LVLMs have achieved strong performance across a wide range of multimodal tasks, including visual question answering, image captioning, and multimodal reasoning~\cite{sima2024drivelm,hartsock2024vision,kim2024image,yang2023exploring,xu2024pllava}. However, despite these advances, they still suffer from hallucination, namely generating outputs that are fluent and plausible at the language level but inconsistent with the actual visual evidence~\cite{bai2024hallucination,favero2024multi,liu2024survey}.
 
\subsection{Contrastive Decoding Calibration}
Contrastive decoding has emerged as one of the most widely-used training-free strategies for mitigating hallucination. Its core idea is to construct a hallucinated branch and calibrate the original prediction by subtracting the hallucinated logits from the orginal logits during decoding~\cite{suo2025octopus,chuang2023dola,leng2024mitigating,wang2024mitigating,manevich2024mitigating,wang2025mitigating,park2025convis,kim2024code,sennrich2024mitigating,wang2024strengthening}. For example, DoLa~\cite{chuang2023dola} first demonstrated in LLMs that contrasting late-layer logits against early-layer logits can improve factuality by surfacing more reliable knowledge. In the multimodal setting, VCD~\cite{leng2024mitigating} contrasts the output distributions conditioned on the original image and a distorted image, thereby suppressing hallucinations caused by excessive reliance on language priors. Subsequent work extends this paradigm by constructing alternative reference distributions. ICD~\cite{wang2024mitigating} introduces disturbance instructions and contrasts them with normal instructions to reveal alignment instability during decoding. LCD~\cite{manevich2024mitigating} instead uses the output distribution of the underlying LLM as a language-only reference to reduce the influence of unsupported visual content. IFCD~\cite{wang2025mitigating} further moves the contrast into the model’s internal space by perturbing hidden representations to simulate hallucination-prone decoding states.

\begin{figure*}[t]
  \centering
  \includegraphics[width=0.9\textwidth]{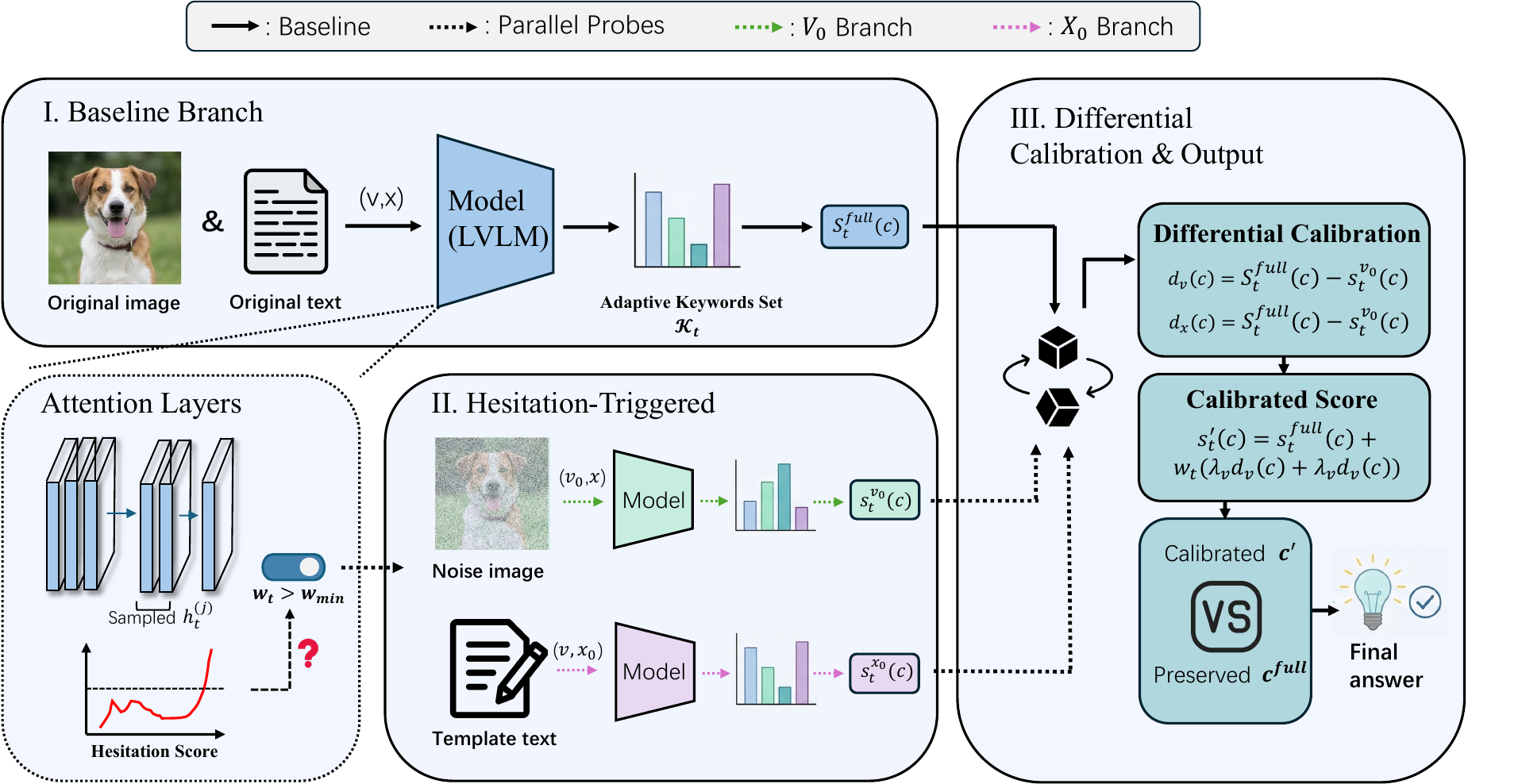}
\caption{Overview of \ours. At each decoding step, the model first performs standard full-branch inference and estimates a hesitation weight $w_t$ from intermediate-layer dynamics. When $w_t > w_{\min}$, two auxiliary probes are activated: a visual perturbation probe ($V0$) and a semantic nullification probe ($X0$). Their score differences with respect to the full branch are then used for differential calibration, yielding the final prediction. A hysteresis margin $\epsilon$ is further applied to avoid unstable switching.}
  \label{fig:overview}
  \Description{}
\end{figure*}

\subsection{Layer-Wise Dynamics for Refinement}

Another line of work mitigates hallucination by exploiting the internal layer-wise dynamics of LVLMs~\cite{wang2025damo, wang2024mllm, tang2025mitigating, jiang2025devils, yu2025hallurnn, liu2025reducing, wu2025sharp}. Its core idea is to analyze how visually grounded evidence evolves across depth and refine decoding with earlier or intermediate representations when later layers become increasingly language-dominated. DAMO~\cite{wang2025damo} shows that hallucination is often associated with localized surges in the late layers and smooths the decoding trajectory with activation momentum. Subsequent work extends this paradigm by using preceding layers more explicitly. DeCo~\cite{wang2024mllm} adaptively incorporates informative earlier layers into the final prediction, while DCLA~\cite{tang2025mitigating} aggregates preceding-layer activations as a dynamic semantic reference to correct semantically deviated layers. Devils in Middle Layers~\cite{jiang2025devils} further shows that the middle layers contain critical stages of visual enrichment and semantic refinement, and leverages attention-lens signals to rebalance visual attention during inference. HalluRNN~\cite{yu2025hallurnn} further introduces recurrent cross-layer reasoning to stabilize information flow across depth. These type of methods are appealing because they are more mechanistically informative: rather than only correcting the final-token distribution, they also help reveal where hallucination forms inside the network. However, many of them directly modify hidden states during decoding, making the intervention relatively intrusive and potentially perturbing the model's native generation dynamics.

\section{Method}
\label{sec:method}

\subsection{Preliminaries}
Let $f_\theta$ denote an LVLM with vocabulary $\mathcal{V}$. At decoding step $t$, given an image $v$, a text query $x$, and previously generated tokens $y_{<t}$, the model produces logits $z_t \in \mathbb{R}^{|\mathcal{V}|}$ and the next-token distribution
\begin{equation}
p_\theta(y_t \mid v, x, y_{<t}) = \operatorname{softmax}(z_t).
\label{eq:next_token}
\end{equation}

For the benchmark settings considered in this work, we further operate on a compact candidate set
\begin{equation}
\mathcal{C} = \{c_1, \ldots, c_M\},
\end{equation}
where each candidate $c \in \mathcal{C}$ is associated with a token set $\mathcal{T}(c)$ containing its possible surface forms. We assign a score to each candidate by aggregating the log-probabilities of the tokens in $\mathcal{T}(c)$:
\begin{equation}
s_t(c) = \operatorname{Agg}
\left(
\left\{
\log p_\theta(\tau \mid v, x, y_{<t}) : \tau \in \mathcal{T}(c)
\right\}
\right),
\label{eq:candidate_score}
\end{equation}
where $\operatorname{Agg}$ is \textsc{LogSumExp} by default, and \textsc{Max} is used in a few task-specific cases.

This candidate-level formulation provides a lightweight and unified interface for calibration in our current setting. For simplicity, we omit the step index $t$ when it is clear from context.

\subsection{Hesitation-Triggered Differential Calibration}
Figure~\ref{fig:overview} shows the overall pipeline of \ours. At decoding step $t$, we first run the \emph{full branch} on the original image-query pair $(v, x)$ under the standard inference process, and obtain the corresponding candidate scores $s_t^{\mathrm{full}}(c)$ for $c \in \mathcal{C}$ via Equation~\ref{eq:candidate_score}. Based on the model's intermediate-layer dynamics, we then compute a hesitation weight $w_t \in [0,1]$, which determines whether the current decoding step requires calibration.

If the model exhibits low hesitation, we keep the full-branch prediction unchanged. Otherwise, \ours activates two lightweight auxiliary probe branches to assess the source of the current preference from two complementary perspectives. The first is a \emph{visual perturbation probe} ($V0$), which weakens the visual evidence while keeping the text input fixed. The second is a \emph{semantic nullification probe} ($X0$), which preserves the image but removes the task-specific semantics of the query while maintaining the answer format. The resulting candidate scores from the two probes are denoted by $s_t^{v0}(c)$ and $s_t^{x0}(c)$, respectively.

These two probes provide complementary counterfactual references for the current prediction. The $V0$ branch reveals how strongly a candidate depends on visual evidence: if the score of a candidate drops substantially after visual perturbation, the candidate is more likely to be visually grounded. In contrast, the $X0$ branch reveals how strongly a candidate depends on the actual query semantics rather than generic response priors: if a candidate remains highly scored even after the semantic content of the query is removed, it is more likely to be supported by superficial answer bias or language inertia.

Based on these probe branches, we define two differential signals
\begin{equation}
\left\{
\begin{aligned}
d_t^v(c) &= s_t^{\mathrm{full}}(c) - s_t^{v0}(c), \\
d_t^x(c) &= s_t^{\mathrm{full}}(c) - s_t^{x0}(c).
\end{aligned}
\right.
\label{eq:dv_dx}
\end{equation}
which measure the change in support for candidate $c$ when visual evidence or query semantics is weakened. Intuitively, a larger $d_t^v(c)$ indicates that the candidate relies more heavily on visual grounding, while a larger $d_t^x(c)$ indicates that it is more strongly supported by the actual query semantics beyond format-level priors.

We then combine these differential signals with the original full-branch score to obtain the calibrated score
\begin{equation}
s_t'(c) = s_t^{\mathrm{full}}(c) + w_t \big( \lambda_v d_t^v(c) + \lambda_x d_t^x(c) \big),
\label{eq:final_calibration}
\end{equation}
where $\lambda_v$ and $\lambda_x$ are scalar coefficients that control the strength of the two calibration terms. In this way, calibration is applied adaptively: when hesitation is low, $w_t$ remains small and the prediction stays close to the original full-branch output; when hesitation is high, the probe-based differential signals are emphasized to revise potentially hallucination-prone candidates. To mitigate the "garbage token" issue, where improbable tokens receive spuriously high scores, we apply an Adaptive Plausibility Constraint (APC). APC restricts the candidate space to the top-$K$ (e.g., $K=200$) tokens of the full-branch distribution, masking all others to $-\infty$ to preserve semantic plausibility.

Finally, \ours selects the prediction according to the calibrated candidate scores. To avoid unstable switching caused by minor score fluctuations, we further apply a small hysteresis margin when comparing the calibrated winner with the original full-branch winner. In the following subsection, we describe how the hesitation weight $w_t$ is computed from intermediate-layer dynamics.

\subsection{Layer-Wise Hesitation as a Trigger Signal}
We now describe the hesitation signal used to determine whether calibration should be activated at decoding step $t$. Our key observation is that hallucination-prone predictions often exhibit unstable intermediate-layer dynamics. When the model is confident and well grounded, its preference over plausible candidates usually evolves in a relatively consistent direction across layers. In contrast, when the model is uncertain, this preference tends to fluctuate rather than converge steadily. We refer to this phenomenon as \emph{layer-wise hesitation}.

Let $h_t^{(j)}$ denote the hidden state of the last token at layer $j$. Using the shared LM head $\phi(\cdot)$, we project each sampled layer to an intermediate logit vector
\begin{equation}
z_t^{(j)} = \phi(h_t^{(j)}).
\label{eq:intermediate_logits}
\end{equation}

\begin{table*}[!t]
  \centering
  \caption{Experimental results on the CHAIR benchmark.}
  \label{tab:chair_results}
  \setlength{\tabcolsep}{12pt}
  \small
  \begin{NiceTabular}{lcccccc}[colortbl-like, cell-space-limits=3pt]
    \toprule
    \rowcolor{headergray}
    \Block{2-1}{\textbf{Method}} 
      & \multicolumn{3}{c}{\textbf{LLaVA-v1.5}} 
      & \multicolumn{3}{c}{\textbf{Qwen3-VL}} \\    
    \rowcolor{headergray}
      & \textbf{CHAIRs} ($\downarrow$) 
      & \textbf{CHAIRi} ($\downarrow$) 
      & \textbf{Recall} ($\uparrow$)
      & \textbf{CHAIRs} ($\downarrow$) 
      & \textbf{CHAIRi} ($\downarrow$) 
      & \textbf{Recall} ($\uparrow$) \\   
    \cmidrule(lr){2-4} \cmidrule(lr){5-7} 
    Regular & 18.2 & 6.2 & 61.9 & 21.6 & 7.8 & \textbf{46.8} \\
    VCD     & 12.2 & 14.3 & 26.5 & 9.2 & 3.0 & 43.4 \\
    ICD     & 22.4 & 8.5 & 55.1 & 6.0 & 3.0 & 29.1 \\
    DoLa    & 16.6 & 9.8 & 55.8 & 4.2 & 5.3 & 29.0 \\
    DAMO    & 17.4 & 6.1 & \textbf{63.1} & \textbf{3.4} & 1.8 & 39.6 \\    
    \midrule
    \rowcolor{rowblue}
    \textbf{Ours} & \textbf{11.6} & \textbf{4.7} & 58.2 & 4.4 & \textbf{1.7} & 39.4 \\
    \bottomrule
  \end{NiceTabular}
\end{table*}

\vspace{1em}
\begin{table}[!t]
  \caption{Experimental results on MME benchmark.}
  \small
  \label{tab:mme}
    \begin{tabular}{llcc}
    \toprule
    \rowcolor{headergray}
    Model & Method & Perception(Total) $\uparrow$ & Cognition (Total) $\uparrow$ \\
    \midrule
    \multirow{6}{*}{LLaVA-v1.5} 
          & Regular & 1508.38 & 357.86 \\
          & VCD     & 1477.74 & 357.86 \\
          & ICD     & 1441.19 & 333.93 \\
          & DoLa    & 1412.13  & 298.57 \\
          & DAMO    & 1512.01 & 297.86 \\
          \rowcolor{rowblue}
          & Ours  & \textbf{1515.89} & \textbf{367.50} \\
    \midrule
    \multirow{6}{*}{Qwen3-VL-4B} 
          & regular & 1701.93 & \textbf{630.71} \\
          & VCD     & 1695.82 & 607.86 \\
          & ICD     & 1660.76 & 597.14 \\
          & DoLa    & 1640.93 & 601.79 \\
          & DAMO    & 1663.72 & 628.57 \\
          \rowcolor{rowblue}
          & Ours  & \textbf{1711.44} & 610.00 \\
    \bottomrule
    \end{tabular}%
\end{table}

To avoid monitoring the entire vocabulary, we construct a compact keyword set $\mathcal{K}_t$ by combining the top-$k$ tokens from the final-layer logits with the token IDs associated with the candidate set:
\begin{equation}
\mathcal{K}_t = \operatorname{TopK}(z_t, k)\ \cup\ \Big( \bigcup_{c \in \mathcal{C}} \mathcal{T}(c) \Big).
\label{eq:keyword_set}
\end{equation}
For each sampled layer, we then compute a keyword-restricted distribution
\begin{equation}
q_t^{(j)} = \operatorname{softmax}\!\big(z_t^{(j)}[\mathcal{K}_t] / \tau_{\mathrm{kw}}\big),
\label{eq:keyword_distribution}
\end{equation}
where $\tau_{\mathrm{kw}}$ is a temperature parameter.

A large change between adjacent layers does not necessarily imply hesitation, since confident refinement may also produce noticeable updates. Instead, we measure whether the current update is consistent with the recent trend. Specifically, we define the layer-to-layer update
\begin{equation}
\Delta q_t^{(j)} = q_t^{(j)} - q_t^{(j-1)},
\label{eq:layer_diff}
\end{equation}
and maintain its exponential moving average
\begin{equation}
\widehat{\Delta q}_t^{(j)} = \alpha \widehat{\Delta q}_t^{(j-1)} + (1-\alpha)\Delta q_t^{(j)},
\label{eq:ema_diff}
\end{equation}
where $\alpha \in [0,1)$ controls the smoothing factor.

We then quantify the hesitation at layer $j$ by the cosine distance between the current update and its smoothed momentum:
\begin{equation}
r_t^{(j)} = 1 - \cos\!\big(\Delta q_t^{(j)}, \widehat{\Delta q}_t^{(j)}\big).
\label{eq:hesitation_per_layer}
\end{equation}
When the two directions are aligned, the model is refining its prediction consistently and the hesitation is small. When they differ substantially, the model is wavering among competing candidates, leading to a larger hesitation value.

Finally, to capture both the overall hesitation trend and sudden layer-wise fluctuations, we define the scalar hesitation score $\mathrm{hes}_t$ by combining the average hesitation (core) and the frequency of high-hesitation spikes across the sampled late layers $\mathcal{J}$:
\begin{equation}
\mathrm{hes}_t = \gamma \cdot \underbrace{\frac{1}{|\mathcal{J}|} \sum_{j \in \mathcal{J}} r_t^{(j)}}_{\text{core}} + \underbrace{\frac{1}{|\mathcal{J}|} \sum_{j \in \mathcal{J}} \mathbb{I}(r_t^{(j)} > \tau_{\text{spike}})}_{\text{spike}},
\label{eq:hesitation_score}
\end{equation}
where $\mathbb{I}(\cdot)$ is the indicator function, $\tau_{\text{spike}}$ is a pre-defined spike threshold, and $\gamma$ is a balancing coefficient. We then convert this score into a soft trigger weight
\begin{equation}
w_t = \sigma\!\left( \frac{\mathrm{hes}_t - \delta}{T} \right),
\label{eq:gate_weight}
\end{equation}
where $\delta$ controls the activation center, $T$ controls the sharpness of the transition, and $\sigma(\cdot)$ is the sigmoid function.

Calibration is activated only when $w_t$ exceeds a preset threshold. In this way, \ours intervenes only at decoding steps that exhibit substantial internal hesitation.

\subsection{Complexity}
\ours incurs low additional overhead by activating auxiliary probes only on hesitation-prone decoding steps. At each step, the full branch is evaluated once as in standard inference. The two probe branches, $V0$ and $X0$, are invoked only on a fraction $r$ of decoding steps satisfying $w_t > w_{\min}$. As a result, the expected number of forward passes per step is
$1 + 2r$, which is substantially smaller than that of methods that apply contrastive calibration at every decoding step. Moreover, the hesitation estimator operates on a compact keyword set and a sparse set of sampled layers, introducing only limited extra computation.

\section{Experiments}

\subsection{Experimental Settings}

\paragraph{\textbf{Benchmarks.}}
We evaluate \ours{} on four representative benchmarks covering both discriminative hallucination detection and open-ended multimodal generation. 
\textbf{MME}~\cite{fu2023mme} is a comprehensive benchmark for multimodal large language models. Following prior hallucination work, we focus on the perception-related split, which is more directly associated with visual grounding, and report the official MME score based on the sum of \textit{Perception} and \textit{Cognition}. 
\textbf{POPE} \cite{li2023evaluating} is designed to evaluate object hallucination through polling-based yes/no probing. We use both the MSCOCO and GQA variants under the random, popular, and adversarial settings, and report \textit{Accuracy} and \textit{F1}, with primary attention to Accuracy and F1. 
\textbf{CHAIR} \cite{rohrbach2018object} evaluates hallucination in open-ended image captioning. Following the standard protocol, we report sentence-level hallucination rate (\textit{CHAIR$_S$}), instance-level hallucination rate (\textit{CHAIR$_I$}) and  the percentage of recalled gt objects over all gt objects (\textit{Recall}). 
\textbf{MM-Vet} \cite{yu2023mm} is a challenging open-ended benchmark for integrated multimodal capabilities. We include it to assess whether hallucination mitigation can preserve broad visual reasoning and instruction-following ability, and report the official overall score.

\paragraph{\textbf{Models.}}
We conduct experiments on two representative open-source LVLM families: \textbf{LLaVA-v1.5} \cite{liu2024improved} and \textbf{Qwen3-VL}  \cite{bai2025qwen3}. These two backbones differ in architecture and multimodal alignment strategy, providing a useful testbed for evaluating the robustness and transferability of \ours{} across models. Unless otherwise specified, we use the default instruction-tuned checkpoints and the official conversation templates released for each model.

\paragraph{\textbf{Baselines.}}
We compare \ours{} with four representative training-free hallucination mitigation baselines, together with the standard decoding of each backbone. 
\textbf{VCD} \cite{leng2024mitigating} mitigates hallucination by contrasting the output distributions from the original image and a visually perturbed image. 
\textbf{ICD} \cite{wang2024mitigating} performs instruction-level contrastive decoding by comparing predictions under the original instruction and a deliberately perturbed instruction. 
\textbf{DoLa}~\cite{chuang2023dola} improves generation by contrasting mature and premature layers during decoding. 
\textbf{DAMO}~\cite{wang2025damo} refines decoding with layer-wise activation momentum to suppress hallucination emerging in later layers. 
These baselines cover input-level contrastive decoding, layer-level contrastive decoding, and hidden-state refinement, providing strong and diverse points of comparison.

\begin{table*}[!ht]
  \centering
  \caption{Experimental results on POPE benchmark.}
  \label{tab:pope}
  \setlength{\tabcolsep}{10pt}
  \begin{NiceTabular}{lllcccccc}[colortbl-like, cell-space-limits=2pt]
    \toprule
    \rowcolor{headergray}
    \Block{2-1}{Dataset} & \Block{2-1}{Model} & \Block{2-1}{Method} & \multicolumn{2}{c}{Random} & \multicolumn{2}{c}{Popular} & \multicolumn{2}{c}{Adversarial} \\
    \rowcolor{headergray}
              &           &       & Acc$\uparrow$ & F1$\uparrow$ & Acc$\uparrow$ & F1$\uparrow$ & Acc$\uparrow$ & F1$\uparrow$ \\
    \cmidrule(lr){4-5} \cmidrule(lr){6-7} \cmidrule(lr){8-9}
    
    \midrule
    \Block[fill=white]{12-1}{MSCOCO} & \Block[fill=white]{6-1}{LLaVA-v1.5} 
              & Regular & 0.8804 & 0.8718 & \textbf{0.8856} & \textbf{0.8756} & 0.8507 & 0.8413\\
              & & VCD     & 0.8818 & 0.8734 & 0.8727 & 0.8614 & 0.8513 & 0.8418 \\
              & & ICD     & 0.8433 & 0.8330 & 0.8607 & 0.8487 & 0.8433 & 0.8330 \\
              & & DoLa    & 0.8821 & 0.8737 & 0.8790 & 0.8736 & 0.8403 & 0.8396 \\
              & & DAMO    & 0.8687 & 0.8564 & 0.8603 & 0.8446 & 0.8433 & 0.8290   \\
              \rowcolor{rowblue}
              & & Ours    & \textbf{0.8825} & \textbf{0.8742} & 0.8730 & 0.8618 & \textbf{0.8523} & \textbf{0.8429} \\
    \cmidrule(lr){2-9} 
              & \Block[fill=white]{6-1}{Qwen3-VL-4B} 
              & Regular & 0.9103 & 0.9058 & 0.8950 & 0.8885 & 0.8847 & 0.8789 \\
              & & VCD     & \textbf{0.9124} & \textbf{0.9081} & 0.8976 & 0.8914 & 0.8860 & 0.8805 \\
              & & ICD     & 0.8733 &  0.8615 & 0.8810 & 0.8687 & \textbf{0.8866} & 0.8775      \\      & & DoLa   & 0.9058 & 0.9040 & 0.8887 & 0.8862 & 0.8760 & 0.8759 \\
              & & DAMO    & 0.9017 & 0.8957 & 0.8903 & 0.8819 & 0.8833 & 0.8753 \\
              \rowcolor{rowblue}
              & & Ours    & \textbf{0.9124} & \textbf{0.9081} & \textbf{0.8977} & \textbf{0.8914} & 0.8860 & \textbf{0.8805} \\ 
    \midrule
    \Block[fill=white]{12-1}{GQA} & \Block[fill=white]{6-1}{LLaVA-v1.5} 
              & Regular   & 0.8680 & 0.8809 & 0.7447 & 0.7926 & 0.6833 & 0.7550 \\
              & & VCD     & 0.7757 & 0.8147 & 0.6453 & 0.8147 & 0.6203 & 0.7220 \\
              & & ICD     & 0.8670 & 0.8800 & 0.7440 & 0.7923 & 0.7313 & 0.7490 \\
              & & DoLa    & \textbf{0.8970} & \textbf{0.9026} & 0.7900 & 0.8196 & 0.7393 & 0.7854 \\
              & & DAMO    & 0.8823 & 0.8913 & 0.7690 & 0.8068 & 0.7090 & 0.7683 \\
              \rowcolor{rowblue}
              & & Ours    & 0.8913 & 0.8967 & \textbf{0.7947} & \textbf{0.8211} & \textbf{0.7430} & \textbf{0.7861} \\
    \cmidrule(lr){2-9}
              & \Block[fill=white]{6-1}{Qwen3-VL-4B} 
              & Regular & 0.9267 & 0.9247 & 0.8803 & 0.8822 & 0.8430 & 0.8517 \\
              & & VCD & \textbf{0.9293} & 0.9193 & 0.8750 & 0.8798 & 0.8317 & 0.8449 \\
              & & ICD & 0.9053 & 0.8990 & 0.8797 & 0.8750 & 0.8457 & 0.8452 \\
              & & DoLa & 0.9210 & 0.9253 & 0.8710 & 0.8850  & 0.8217 & 0.8522 \\
              & & DAMO & 0.9260 & 0.9241 & 0.8800 & 0.8820 & 0.8423 & 0.8513 \\
              \rowcolor{rowblue}
              & & Ours & 0.9287 & \textbf{0.9272} & \textbf{0.8847} & \textbf{0.8873}  & \textbf{0.8470} & \textbf{0.8544} \\
    \bottomrule
  \end{NiceTabular}
\end{table*}

\begin{table}[t]
\centering
\caption{Computational overhead comparison on the MME benchmark}
\label{tab:overhead}
\renewcommand{\arraystretch}{1.3}
\resizebox{\linewidth}{!}{
\begin{NiceTabular}{llcccc}[colortbl-like]
\toprule
\rowcolor{headergray}
\textbf{Method} & \textbf{Intervention} & \textbf{Trigger Rate} & \textbf{$N_{fwd}$} & \textbf{Latency (ms)} & \textbf{Overhead} \\
\midrule
Regular  & None  & 0.0\%  & 1.000  & 60.77  & - \\
\midrule
VCD  & Logit (Continuous)  & 100.0\%  & 2.000  & 168.80  & +177.8\% \\
ICD  & Logit (Continuous)  & 100.0\%  & 2.000  & 169.75  & +179.3\% \\
\rowcolor{rowblue}
\textbf{Ours}  & \textbf{Logit (Dynamic)} & \textbf{5.3\%}   & \textbf{1.107}     & \textbf{71.08}        & \textbf{+17.0\%} \\
\bottomrule
\end{NiceTabular}
}
\end{table}

\vspace{-1mm}
\paragraph{\textbf{Implementation Details.}}
All methods are evaluated in a strictly training-free setting, without any additional finetuning or external supervision. To ensure fair comparison, we keep the prompt format, image preprocessing pipeline, and decoding configuration identical across methods for each backbone. For discriminative benchmarks such as MME and POPE, we use deterministic decoding with temperature set to 0 and top-$p$ set to 1, and evaluate candidate-level predictions under the official benchmark protocols. For open-ended benchmarks such as CHAIR and MM-Vet, we follow the standard generation and evaluation pipelines provided by the corresponding benchmarks. For VCD, ICD, DoLa, and DAMO, we use their official implementations when available, or otherwise follow the hyperparameter settings reported in the original papers. For \ours{}, hyperparameters including the sampled layers, keyword set size, hesitation threshold, triggering gate, and calibration coefficients are tuned on a held-out validation split and then fixed across all reported experiments. No retrieval module, verifier model, or post-hoc correction stage is introduced.

\subsection{Experimental Results}

\begin{table*}[t]
    \centering
    \caption{Experimental results on MM-Vet benchmark. We report the performance across different capability dimensions: recognition (rec), OCR (ocr), knowledge (know), generation (gen), spatial (spat), and math (math).}
    \label{tab:mm-vet}
    \setlength{\tabcolsep}{12pt}
    \begin{NiceTabular}{llccccccc}[colortbl-like, cell-space-limits=2pt]
        \toprule
        \rowcolor{headergray}
        Model & Method & rec ($\uparrow$) & ocr ($\uparrow$) & know ($\uparrow$) & gen ($\uparrow$) & spat ($\uparrow$) & math ($\uparrow$) & Total ($\uparrow$) \\
        \midrule
        
        \Block[fill=white]{6-1}{LLaVA-v1.5} & Regular & \underline{34.4} & \underline{23.0} & 16.1 & 20.5 & 25.2 & \underline{11.5} & 29.9 \\
                                        & VCD     & 7.5 & 8.9 & 10.1 & 14.5 & 6.3 & 5.8 & 7.1 \\
                                        & ICD     & 27.0   & 20.1   & 15.9   & \underline{20.6}   & 24.6   & 7.8   & 29.8   \\
                                        & DoLa    & 33.2   & 22.6   & 17.3   & 13.0   & \textbf{28.1}   & \textbf{18.5}   & \textbf{30.7}   \\
                                        & DAMO    & 33.5   & 20.4   & \textbf{18.0}   & 19.7   & 23.7   & 7.7   & 29.3 \\
        \rowcolor{rowblue}              & Ours    & \textbf{35.1} & \textbf{23.2} & \underline{17.4} & \textbf{21.0} & \underline{25.5} & \underline{11.5} & \textbf{30.7} \\
        
        \midrule
        
        \Block[fill=white]{6-1}{Qwen3-vl}   & Regular & \textbf{64.2}   & 77.9   & 50.8   & 51.6   & 76.4   & \underline{80.0}   & \underline{69.0}   \\
                                        & VCD     & 63.9   & \underline{80.3}   & 50.8   & 51.1   & \underline{79.3}   & 75.8   & \underline{69.0}   \\
                                        & ICD     & 63.5   & 79.3   & \textbf{52.9}   & \textbf{53.1}   & 77.6   & 75.4   & \underline{69.0}   \\
                                        & DoLa    & 54.3   & 65.4   & 38.1   & 38.1   & 68.7   & 49.2   & 58.2   \\
                                        & DAMO    & 54.8   & 58.6   & 39.6   & 37.1   & 61.1   & 48.1   & 56.7   \\
        \rowcolor{rowblue}              & Ours    & 
        \underline{64.0} & \textbf{82.7} & \underline{51.9} & \underline{51.9} & \textbf{80.4} & \textbf{82.3} & \textbf{70.3} \\
        
        \bottomrule
    \end{NiceTabular}
\end{table*}

\paragraph{\textbf{Results on MME}}
Table~\ref{tab:mme} shows that HTDC is particularly effective on the grounding-sensitive MME benchmark. On LLaVA-v1.5, it achieves the best scores on both \textit{Perception} (1515.89) and \textit{Cognition} (367.50), improving over Regular decoding by +7.51 and +9.64, respectively. On Qwen3-VL-4B, HTDC further obtains the best \textit{Perception} score (1711.44), outperforming Regular, VCD, ICD, and DAMO, while its \textit{Cognition} score (610.00) remains higher than VCD,ICD and DoLa but below Regular and DAMO. Overall, these results suggest that HTDC is particularly effective for grounding-sensitive perception tasks, where selective calibration can strengthen visual grounding without the over-correction introduced by always-on decoding methods.

\vspace{-2mm}
\paragraph{\textbf{Results on POPE}}
Table~\ref{tab:pope} shows that \ours performs strongly on POPE, with the clearest gains on the more challenging GQA setting. On MSCOCO, HTDC achieves the best results on LLaVA-v1.5 Random and Adversarial, and remains highly competitive across all subsets on Qwen3-VL-4B. On GQA, \ours further delivers the best results on Popular and Adversarial for both backbones, while remaining near-best on Random. These results suggest that HTDC is especially effective when object hallucination is driven by stronger language inertia and co-occurrence bias, where hesitation-triggered calibration can more accurately suppress spurious affirmative predictions without unnecessarily perturbing stable decoding steps.

\vspace{-1mm}
\paragraph{\textbf{Results on CHAIR}}
As shown in Table~\ref{tab:chair_results}, \ours is particularly effective on CHAIR. On LLaVA-v1.5, it achieves the best hallucination suppression, reducing \textit{CHAIR}$_S$ from 18.2 to 11.6 and \textit{CHAIR}$_I$ from 6.2 to 4.7, while maintaining competitive Recall (58.2). This indicates that hesitation-triggered calibration can suppress hallucinated objects without the severe coverage loss caused by always-on correction, e.g., VCD drops Recall to 26.5. On Qwen3-VL, HTDC remains highly competitive, achieving the best \textit{CHAIR}$_I$ (1.7), near-best \textit{CHAIR}$_S$ (4.4), and Recall comparable to DAMO (39.4 vs.\ 39.6). Overall, these results show that HTDC provides a favorable hallucination--recall trade-off for open-ended generation, with especially strong gains on LLaVA-v1.5.

\paragraph{\textbf{Results on MM-Vet}}
Table~\ref{tab:mm-vet} demonstrates that \ours effectively mitigates hallucinations without degrading general multimodal capabilities. On LLaVA-v1.5, \ours achieves a top Total score of 30.7. Crucially, it completely avoids the catastrophic collapse seen in VCD, whose score plummets to 7.1 due to the continuous over-penalization of fragile reasoning steps. On Qwen3-VL, \ours further improves the baseline performance to 70.3, whereas intrusive internal-state refinement methods like DoLa and DAMO disrupt the model's native representations and suffer significant drops (falling to 58.2 and 56.7, respectively). Overall, these results highlight that our hesitation-triggered mechanism successfully preserves the model's inherent reasoning and instruction-following abilities, achieving an optimal balance between hallucination suppression and general capability.

\begin{figure}[h]
  \centering
  \includegraphics[width=\linewidth]{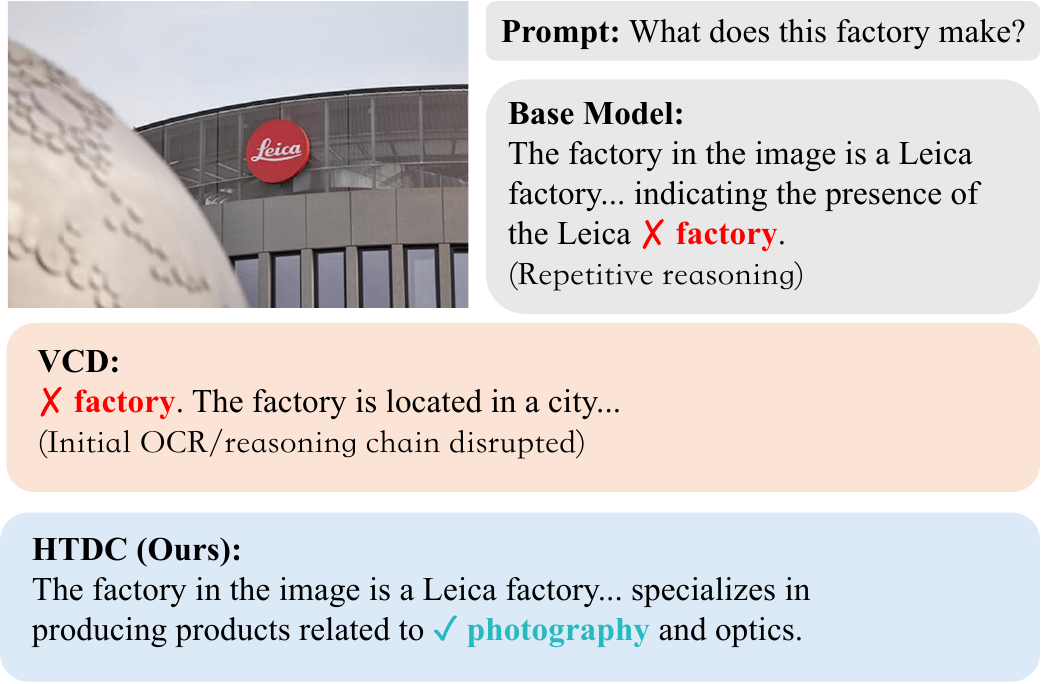}
   \caption{Qualitative comparison on complex reasoning (MM-Vet). VCD's continuous intervention destroys the initial OCR-based reasoning chain, whereas \ours preserves stable generation and accurate reasoning capabilities.}
  \label{fig:case2}
\end{figure}

\paragraph{\textbf{Case Study}}
As shown in Figure~\ref{fig:case2}, we present a case study to further demonstrate the effectiveness of \ours on LLaVA1.5. While the vanilla setting and VCD fails to answer the given question, \ours successfully correct the vanilla and VCD deocoding, leading to accurate response. 

\subsection{Computational Efficiency Analysis}
While continuous contrastive decoding methods (e.g., VCD, ICD) effectively mitigate hallucinations, their reliance on step-wise dual-branch generation imposes a severe computational bottleneck. In autoregressive decoding, computing the contrastive penalty for every single token intrinsically doubles the theoretical forward passes ($N_{fwd}=2.0$). As shown in Table~\ref{tab:overhead}, this continuous intervention drastically inflates the generation latency on MME from 60.77 ms (Regular) to over 168 ms.

Our proposed \ours fundamentally resolves this inefficiency by shifting to a dynamic, hesitation-triggered paradigm. By leveraging internal epistemic uncertainty as a lightweight vanguard signal, HTDC only activates the computationally heavy visual ($V_0$) and semantic ($X_0$) probes when the LVLM is prone to hallucinatory generation. Across the MME evaluation, HTDC maintains an extremely sparse trigger rate of 5.3\%, compressing the expected forward passes per token ($N_{fwd}$) to just 1.107. Ultimately, HTDC achieves robust differential calibration with a marginal latency overhead of merely +17.0\% (71.08 ms per token). This stark contrast (+17.0\% vs. +180\%) demonstrates that HTDC successfully breaks the persistent trade-off between decoding speed and hallucination mitigation, making advanced contrastive decoding highly feasible for real-time deployment.

\begin{figure*}[t]
  \centering
  \includegraphics[width=0.95\textwidth]{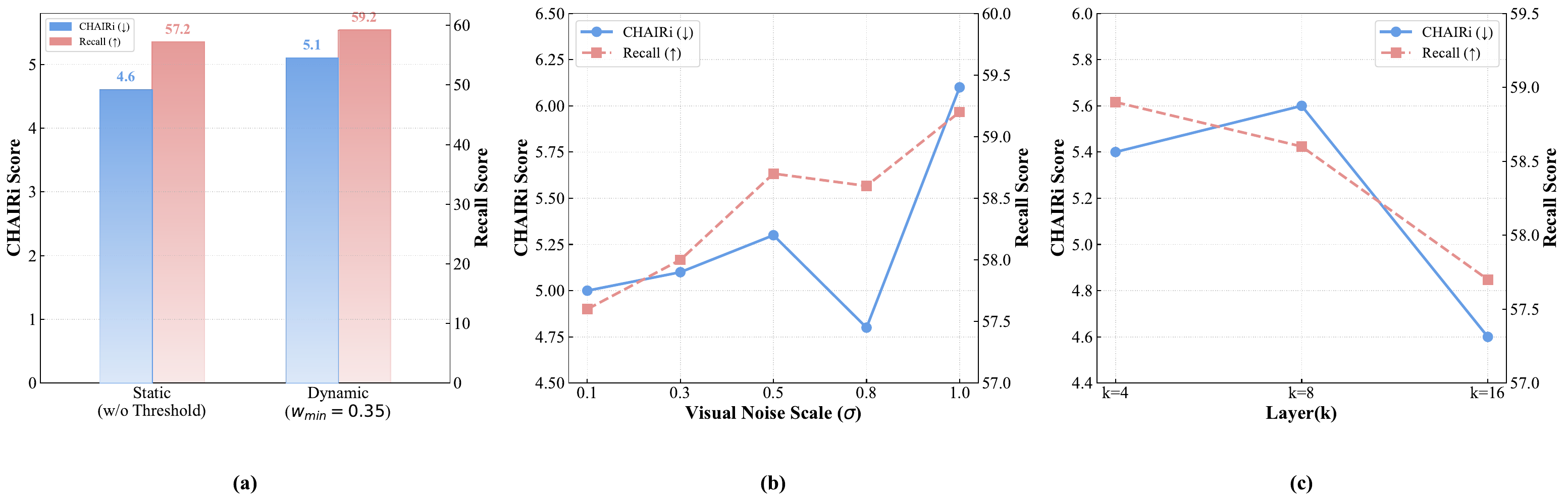}
  \caption{Impact of key hyperparameters on the CHAIR benchmark, including (a) dynamic vs. static gating strategy, (b) visual noise scale ($\sigma$), and (c) layer selection depth ($k$).}
  \label{fig:ablations}
  \Description{Line and bar charts demonstrating that dynamic gating recovers recall, noise scale at 0.8 optimizes CHAIRi, and layer depth adjustments trade off between recall and hallucination suppression.}
\end{figure*}

\section{Ablation Studies}
To comprehensively understand the working mechanisms of our \ours framework, we conduct systematic ablation studies on its core components and key hyperparameters.

\subsection{Synergy of Differential Calibration Branches}
Table~\ref{tab:ablation_components} decouples the effects of the visual branch ($V_0$) and the semantic branch ($X_0$). An intriguing task-specific phenomenon is observed:

\begin{itemize}
    \item \textbf{Discriminative Tasks (POPE):} Removing the visual branch yields the highest performance (87.26\% Acc). In Yes/No evaluations, the primary source of error is the LLM's statistical bias toward affirmative answers. The semantic branch ($X_0$) serves as a targeted remedy for this ``language inertia,'' precisely stripping the directive bias without disturbing fine-grained pixel evidence.
    \item \textbf{Generative Tasks (CHAIR):} In contrast, the visual branch ($V_0$) is indispensable for open-ended captioning. Stripping the semantic branch reveals that the visual perturbation explicitly forces the model to mine deeper visual features, driving Recall to a peak of 62.2, albeit at the cost of slightly higher hallucinations.
\end{itemize}

Ultimately, the Full \ours orchestrates these orthogonal strengths, delivering state-of-the-art hallucination suppression ($\text{CHAIR}_I$ 4.7) while generalizing powerfully across diverse evaluation paradigms.

\subsection{Hyperparameter Sensitivity}
Figure~\ref{fig:ablations} illustrates the impact of critical control mechanisms within our framework, evaluated on the sensitive CHAIR benchmark.

\begin{itemize}
    \item \textbf{Gating Strategy (Figure~\ref{fig:ablations}(a)):} Applying static calibration relentlessly penalizes every token, which minimizes hallucinations but severely harms the Recall rate (57.2). Our dynamic hesitation-triggered gating successfully preserves confident, accurate generations, recovering Recall to 59.2.
    \item \textbf{Visual Noise Scale (Figure~\ref{fig:ablations}(b)):} A non-monotonic trend is observed across varying noise scales ($\sigma$). A moderate noise level ($\sigma=0.8$) optimally exposes visual vulnerabilities (achieving the lowest $\text{CHAIR}_I$), whereas excessive noise ($\sigma=1.0$) corrupts the contrastive reference.
    \item \textbf{Layer Selection (Figure~\ref{fig:ablations}(c)):} The probing depth $k$ dictates the sensitivity of the hesitation trigger. A larger $k$ ($k=16$) captures early, minute uncertainties, yielding aggressive hallucination suppression ($\text{CHAIR}_I$ 4.6), while a smaller $k$ ($k=4, 8$) focuses on late-stage logit flips, offering a balanced trade-off.
\end{itemize}

\begin{table}[H]
  \centering
  \caption{Ablation studies on CHAIR and POPE.}
  \label{tab:ablation_components}
  \begin{NiceTabular}{lccccc}[colortbl-like, cell-space-limits=2pt]
    \toprule
    \rowcolor{headergray}
    \Block{2-1}{Method} & \multicolumn{3}{c}{CHAIR} & \multicolumn{2}{c}{POPE} \\
    \cmidrule(lr){2-4} \cmidrule(lr){5-6}
    \rowcolor{headergray}
    & C$_S$$\downarrow$ & C$_I$$\downarrow$ & Recall$\uparrow$ & Acc$\uparrow$ & F1$\uparrow$ \\
    \midrule
    Ours (Full HTDC) & \textbf{11.6} & \textbf{4.7} & 58.2 & 86.94 & 86.91 \\
    w/o visual branch & 11.6 & 4.7 & 58.3 & \textbf{87.26} & \textbf{87.11} \\
    w/o semantic branch & 18.2 & 6.6 & \textbf{62.2} & 86.12 & 85.45 \\
    w/o both (Baseline) & 18.2 & 6.2 & 61.9 & 86.86 & 85.89 \\
    \bottomrule
  \end{NiceTabular}
\end{table}

\section{Discussion and Limitations}
While \ours provides an effective and highly efficient training-free solution for hallucination mitigation, it has a few limitations that warrant future exploration. First, because our layer-wise hesitation metric relies on intermediate hidden states and logit distributions, \ours intrinsically requires white-box access to the model's internal representations. Consequently, it cannot be directly applied to closed-source models accessed via black-box APIs. Second, the optimal hesitation triggering threshold ($\delta$) and the APC cutoff ($K$) may exhibit slight variations across different LVLM families due to their distinct pre-training and alignment dynamics, necessitating a modest calibration phase on a validation set. Finally, our current framework primarily operates at the token level. Extending this hesitation-triggered mechanism to multi-token candidate decoding, or designing more sophisticated multimodal nullification operators for emerging domains like video-language reasoning, represent promising directions for future work.

\section{Conclusion}
In this paper, we proposed Hesitation-Triggered Differential Calibration (HTDC), a novel, training-free decoding framework designed to efficiently mitigate hallucinations in Large Vision-Language Models. By identifying layer-wise hesitation as a reliable internal signal of grounding uncertainty, \ours shifts the hallucination-mitigation paradigm from continuous, computationally heavy intervention to a lightweight, dynamic gating mechanism. When triggered, HTDC adaptively calibrates the prediction using complementary visual and semantic nullification probes, reinforced by an Adaptive Plausibility Constraint (APC) to ensure generation stability. Extensive experiments demonstrate that \ours effectively suppresses object and attribute hallucinations across diverse benchmarks (e.g., POPE, CHAIR, MME) while rigorously preserving the model's general multimodal capabilities on complex tasks like MM-Vet. Crucially, \ours achieves these state-of-the-art results with a marginal latency overhead of merely 17.0\%, successfully breaking the persistent trade-off between decoding speed and hallucination mitigation, and paving the way for the robust real-time deployment of LVLMs.

\bibliographystyle{acmmm}
\bibliography{reference}

@String{Computer = "{IEEE} Computer" }

@String{Chelsea = "Chelsea" }

@String{Springer = "Springer-Verlag" }

@article{lee2018pre,
  title={Pre-training of deep bidirectional transformers for language understanding},
  author={Lee, JDMCK and Toutanova, K},
  journal={arXiv preprint arXiv:1810.04805},
  volume={3},
  number={8},
  pages={4171--4186},
  year={2018}
}

@article{liu2023visual,
  title={Visual instruction tuning},
  author={Liu, Haotian and Li, Chunyuan and Wu, Qingyang and Lee, Yong Jae},
  journal={Advances in neural information processing systems},
  volume={36},
  pages={34892--34916},
  year={2023}
}

@inproceedings{liu2024improved,
  title={Improved baselines with visual instruction tuning},
  author={Liu, Haotian and Li, Chunyuan and Li, Yuheng and Lee, Yong Jae},
  booktitle={Proceedings of the IEEE/CVF conference on computer vision and pattern recognition},
  pages={26296--26306},
  year={2024}
}

@article{wang2024qwen2,
  title={Qwen2-vl: Enhancing vision-language model's perception of the world at any resolution},
  author={Wang, Peng and Bai, Shuai and Tan, Sinan and Wang, Shijie and Fan, Zhihao and Bai, Jinze and Chen, Keqin and Liu, Xuejing and Wang, Jialin and Ge, Wenbin and others},
  journal={arXiv preprint arXiv:2409.12191},
  year={2024}
}

@article{bai2025qwen3,
  title={Qwen3-vl technical report},
  author={Bai, Shuai and Cai, Yuxuan and Chen, Ruizhe and Chen, Keqin and Chen, Xionghui and Cheng, Zesen and Deng, Lianghao and Ding, Wei and Gao, Chang and Ge, Chunjiang and others},
  journal={arXiv preprint arXiv:2511.21631},
  year={2025}
}

@article{yin2024woodpecker,
  title={Woodpecker: Hallucination correction for multimodal large language models},
  author={Yin, Shukang and Fu, Chaoyou and Zhao, Sirui and Xu, Tong and Wang, Hao and Sui, Dianbo and Shen, Yunhang and Li, Ke and Sun, Xing and Chen, Enhong},
  journal={Science China Information Sciences},
  volume={67},
  number={12},
  pages={220105},
  year={2024},
  publisher={Springer}
}

@article{bai2024hallucination,
  title={Hallucination of multimodal large language models: A survey},
  author={Bai, Zechen and Wang, Pichao and Xiao, Tianjun and He, Tong and Han, Zongbo and Zhang, Zheng and Shou, Mike Zheng},
  journal={arXiv preprint arXiv:2404.18930},
  year={2024}
}

@article{yin2024survey,
  title={A survey on multimodal large language models},
  author={Yin, Shukang and Fu, Chaoyou and Zhao, Sirui and Li, Ke and Sun, Xing and Xu, Tong and Chen, Enhong},
  journal={National Science Review},
  volume={11},
  number={12},
  pages={nwae403},
  year={2024},
  publisher={Oxford University Press}
}

@inproceedings{li2024manipllm,
  title={Manipllm: Embodied multimodal large language model for object-centric robotic manipulation},
  author={Li, Xiaoqi and Zhang, Mingxu and Geng, Yiran and Geng, Haoran and Long, Yuxing and Shen, Yan and Zhang, Renrui and Liu, Jiaming and Dong, Hao},
  booktitle={Proceedings of the IEEE/CVF Conference on Computer Vision and Pattern Recognition},
  pages={18061--18070},
  year={2024}
}

@inproceedings{ma2024coco,
  title={Coco-agent: A comprehensive cognitive mllm agent for smartphone gui automation},
  author={Ma, Xinbei and Zhang, Zhuosheng and Zhao, Hai},
  booktitle={Findings of the Association for Computational Linguistics: ACL 2024},
  pages={9097--9110},
  year={2024}
}

@inproceedings{wang2024mitigating,
  title={Mitigating fine-grained hallucination by fine-tuning large vision-language models with caption rewrites},
  author={Wang, Lei and He, Jiabang and Li, Shenshen and Liu, Ning and Lim, Ee-Peng},
  booktitle={International Conference on Multimedia Modeling},
  pages={32--45},
  year={2024},
  organization={Springer}
}

@inproceedings{yu2024rlhf,
  title={Rlhf-v: Towards trustworthy mllms via behavior alignment from fine-grained correctional human feedback},
  author={Yu, Tianyu and Yao, Yuan and Zhang, Haoye and He, Taiwen and Han, Yifeng and Cui, Ganqu and Hu, Jinyi and Liu, Zhiyuan and Zheng, Hai-Tao and Sun, Maosong and others},
  booktitle={Proceedings of the IEEE/CVF Conference on Computer Vision and Pattern Recognition},
  pages={13807--13816},
  year={2024}
}

@inproceedings{yang2025mitigating,
  title={Mitigating hallucinations in large vision-language models via dpo: On-policy data hold the key},
  author={Yang, Zhihe and Luo, Xufang and Han, Dongqi and Xu, Yunjian and Li, Dongsheng},
  booktitle={Proceedings of the IEEE/CVF Conference on Computer Vision and Pattern Recognition},
  pages={10610--10620},
  year={2025}
}

@inproceedings{leng2024mitigating,
  title={Mitigating object hallucinations in large vision-language models through visual contrastive decoding},
  author={Leng, Sicong and Zhang, Hang and Chen, Guanzheng and Li, Xin and Lu, Shijian and Miao, Chunyan and Bing, Lidong},
  booktitle={Proceedings of the IEEE/CVF Conference on Computer Vision and Pattern Recognition},
  pages={13872--13882},
  year={2024}
}

@article{chuang2023dola,
  title={Dola: Decoding by contrasting layers improves factuality in large language models},
  author={Chuang, Yung-Sung and Xie, Yujia and Luo, Hongyin and Kim, Yoon and Glass, James and He, Pengcheng},
  journal={arXiv preprint arXiv:2309.03883},
  year={2023}
}

@inproceedings{wang2025damo,
  title={Damo: Decoding by accumulating activations momentum for mitigating hallucinations in vision-language models},
  author={Wang, Kaishen and Gu, Hengrui and Gao, Meijun and Zhou, Kaixiong},
  booktitle={The Thirteenth International Conference on Learning Representations},
  year={2025}
}

@article{yu2025hallurnn,
  title={Hallurnn: Mitigating hallucinations via recurrent cross-layer reasoning in large vision-language models},
  author={Yu, Le and Wang, Kaishen and Xiong, Jianlong and Cao, Yue and Zhang, Lei and He, Zhang Yi Tao},
  journal={arXiv preprint arXiv:2506.17587},
  year={2025}
}

@inproceedings{manevich2024mitigating,
  title={Mitigating hallucinations in large vision-language models (lvlms) via language-contrastive decoding (lcd)},
  author={Manevich, Avshalom and Tsarfaty, Reut},
  booktitle={Findings of the Association for Computational Linguistics: ACL 2024},
  pages={6008--6022},
  year={2024}
}

@article{wang2025mitigating,
  title={Mitigating Hallucinations in Large Vision-Language Models with Internal Fact-based Contrastive Decoding},
  author={Wang, Chao and Zhou, Xuancheng and Fu, Weiwei and Zhou, Yang},
  journal={arXiv preprint arXiv:2502.01056},
  year={2025}
}

@inproceedings{li2023evaluating,
  title={Evaluating object hallucination in large vision-language models},
  author={Li, Yifan and Du, Yifan and Zhou, Kun and Wang, Jinpeng and Zhao, Wayne Xin and Wen, Ji-Rong},
  booktitle={Proceedings of the 2023 conference on empirical methods in natural language processing},
  pages={292--305},
  year={2023}
}

@article{fu2023mme,
  title={Mme: A comprehensive evaluation benchmark for multimodal large language models},
  author={Fu, Chaoyou and Chen, Peixian and Shen, Yunhang and Qin, Yulei and Zhang, Mengdan and Lin, Xu and Yang, Jinrui and Zheng, Xiawu and Li, Ke and Sun, Xing and others},
  journal={arXiv preprint arXiv:2306.13394},
  year={2023}
}

@inproceedings{rohrbach2018object,
  title={Object hallucination in image captioning},
  author={Rohrbach, Anna and Hendricks, Lisa Anne and Burns, Kaylee and Darrell, Trevor and Saenko, Kate},
  booktitle={Proceedings of the 2018 Conference on Empirical Methods in Natural Language Processing},
  pages={4035--4045},
  year={2018}
}

@article{zhao2024mitigating,
  title={Mitigating object hallucination in large vision-language models via image-grounded guidance},
  author={Zhao, Linxi and Deng, Yihe and Zhang, Weitong and Gu, Quanquan},
  journal={arXiv preprint arXiv:2402.08680},
  year={2024}
}

@inproceedings{favero2024multi,
  title={Multi-modal hallucination control by visual information grounding},
  author={Favero, Alessandro and Zancato, Luca and Trager, Matthew and Choudhary, Siddharth and Perera, Pramuditha and Achille, Alessandro and Swaminathan, Ashwin and Soatto, Stefano},
  booktitle={Proceedings of the IEEE/CVF Conference on Computer Vision and Pattern Recognition},
  pages={14303--14312},
  year={2024}
}

@inproceedings{suo2025octopus,
  title={Octopus: Alleviating hallucination via dynamic contrastive decoding},
  author={Suo, Wei and Zhang, Lijun and Sun, Mengyang and Wu, Lin Yuanbo and Wang, Peng and Zhang, Yanning},
  booktitle={Proceedings of the Computer Vision and Pattern Recognition Conference},
  pages={29904--29914},
  year={2025}
}

@inproceedings{sennrich2024mitigating,
  title={Mitigating hallucinations and off-target machine translation with source-contrastive and language-contrastive decoding},
  author={Sennrich, Rico and Vamvas, Jannis and Mohammadshahi, Alireza},
  booktitle={Proceedings of the 18th Conference of the European Chapter of the Association for Computational Linguistics (Volume 2: Short Papers)},
  pages={21--33},
  year={2024}
}

@article{kim2024code,
  title={Code: Contrasting self-generated description to combat hallucination in large multi-modal models},
  author={Kim, Junho and Kim, Hyun J and Kim, Yeon J and Ro, Yong M},
  journal={Advances in Neural Information Processing Systems},
  volume={37},
  pages={133571--133599},
  year={2024}
}

@article{wang2024strengthening,
  title={Strengthening layer interaction via dynamic layer attention},
  author={Wang, Kaishen and Xia, Xun and Liu, Jian and Yi, Zhang and He, Tao},
  journal={arXiv preprint arXiv:2406.13392},
  year={2024}
}

@inproceedings{sima2024drivelm,
  title={Drivelm: Driving with graph visual question answering},
  author={Sima, Chonghao and Renz, Katrin and Chitta, Kashyap and Chen, Li and Zhang, Hanxue and Xie, Chengen and Bei{\ss}wenger, Jens and Luo, Ping and Geiger, Andreas and Li, Hongyang},
  booktitle={European conference on computer vision},
  pages={256--274},
  year={2024},
  organization={Springer}
}

@article{xu2024pllava,
  title={Pllava: Parameter-free llava extension from images to videos for video dense captioning},
  author={Xu, Lin and Zhao, Yilin and Zhou, Daquan and Lin, Zhijie and Ng, See Kiong and Feng, Jiashi},
  journal={arXiv preprint arXiv:2404.16994},
  year={2024}
}

@article{yang2023exploring,
  title={Exploring diverse in-context configurations for image captioning},
  author={Yang, Xu and Wu, Yongliang and Yang, Mingzhuo and Chen, Haokun and Geng, Xin},
  journal={Advances in Neural Information Processing Systems},
  volume={36},
  pages={40924--40943},
  year={2023}
}

@article{kim2024image,
  title={An image grid can be worth a video: Zero-shot video question answering using a vlm},
  author={Kim, Wonkyun and Choi, Changin and Lee, Wonseok and Rhee, Wonjong},
  journal={IEEE Access},
  volume={12},
  pages={193057--193075},
  year={2024},
  publisher={IEEE}
}

@article{hartsock2024vision,
  title={Vision-language models for medical report generation and visual question answering: A review},
  author={Hartsock, Iryna and Rasool, Ghulam},
  journal={Frontiers in artificial intelligence},
  volume={7},
  pages={1430984},
  year={2024},
  publisher={Frontiers Media SA}
}

@article{zhou2024aligning,
  title={Aligning modalities in vision large language models via preference fine-tuning},
  author={Zhou, Yiyang and Cui, Chenhang and Rafailov, Rafael and Finn, Chelsea and Yao, Huaxiu},
  journal={arXiv preprint arXiv:2402.11411},
  year={2024}
}

@inproceedings{zhang2025active,
  title={Active layer-contrastive decoding reduces hallucination in large language model generation},
  author={Zhang, Hongxiang and Chen, Hao and Chen, Muhao and Zhang, Tianyi},
  booktitle={Proceedings of the 2025 Conference on Empirical Methods in Natural Language Processing},
  pages={3028--3046},
  year={2025}
}

@inproceedings{park2025convis,
  title={Convis: Contrastive decoding with hallucination visualization for mitigating hallucinations in multimodal large language models},
  author={Park, Yeji and Lee, Deokyeong and Choe, Junsuk and Chang, Buru},
  booktitle={Proceedings of the AAAI Conference on Artificial Intelligence},
  volume={39},
  number={6},
  pages={6434--6442},
  year={2025}
}

@article{lee2024delve,
  title={Delve into visual contrastive decoding for hallucination mitigation of large vision-language models},
  author={Lee, Yi-Lun and Tsai, Yi-Hsuan and Chiu, Wei-Chen},
  journal={arXiv preprint arXiv:2412.06775},
  year={2024}
}

@article{mir2025geometry,
  title={The Geometry of Truth: Layer-wise Semantic Dynamics for Hallucination Detection in Large Language Models},
  author={Mir, Amir Hameed},
  journal={arXiv preprint arXiv:2510.04933},
  year={2025}
}

@inproceedings{maynez2020faithfulness,
  title={On faithfulness and factuality in abstractive summarization},
  author={Maynez, Joshua and Narayan, Shashi and Bohnet, Bernd and McDonald, Ryan},
  booktitle={Proceedings of the 58th annual meeting of the association for computational linguistics},
  pages={1906--1919},
  year={2020}
}

@article{huang2025survey,
  title={A survey on hallucination in large language models: Principles, taxonomy, challenges, and open questions},
  author={Huang, Lei and Yu, Weijiang and Ma, Weitao and Zhong, Weihong and Feng, Zhangyin and Wang, Haotian and Chen, Qianglong and Peng, Weihua and Feng, Xiaocheng and Qin, Bing and others},
  journal={ACM Transactions on Information Systems},
  volume={43},
  number={2},
  pages={1--55},
  year={2025},
  publisher={ACM New York, NY}
}

@article{liu2024survey,
  title={A survey on hallucination in large vision-language models},
  author={Liu, Hanchao and Xue, Wenyuan and Chen, Yifei and Chen, Dapeng and Zhao, Xiutian and Wang, Ke and Hou, Liping and Li, Rongjun and Peng, Wei},
  journal={arXiv preprint arXiv:2402.00253},
  year={2024}
}

@article{hager2024evaluation,
  title={Evaluation and mitigation of the limitations of large language models in clinical decision-making},
  author={Hager, Paul and Jungmann, Friederike and Holland, Robbie and Bhagat, Kunal and Hubrecht, Inga and Knauer, Manuel and Vielhauer, Jakob and Makowski, Marcus and Braren, Rickmer and Kaissis, Georgios and others},
  journal={Nature medicine},
  volume={30},
  number={9},
  pages={2613--2622},
  year={2024},
  publisher={Nature Publishing Group US New York}
}

@article{kang2023deficiency,
  title={Deficiency of large language models in finance: An empirical examination of hallucination},
  author={Kang, Haoqiang and Liu, Xiao-Yang},
  journal={arXiv preprint arXiv:2311.15548},
  year={2023}
}

@article{yu2023mm,
  title={Mm-vet: Evaluating large multimodal models for integrated capabilities},
  author={Yu, Weihao and Yang, Zhengyuan and Li, Linjie and Wang, Jianfeng and Lin, Kevin and Liu, Zicheng and Wang, Xinchao and Wang, Lijuan},
  journal={arXiv preprint arXiv:2308.02490},
  year={2023}
}

@article{wang2024mllm,
  title={Mllm can see? dynamic correction decoding for hallucination mitigation},
  author={Wang, Chenxi and Chen, Xiang and Zhang, Ningyu and Tian, Bozhong and Xu, Haoming and Deng, Shumin and Chen, Huajun},
  journal={arXiv preprint arXiv:2410.11779},
  year={2024}
}

@article{tang2025mitigating,
  title={Mitigating Hallucinations via Inter-Layer Consistency Aggregation in Large Vision-Language Models},
  author={Tang, Kai and You, Jinhao and Ge, Xiuqi and Li, Hanze and Guo, Yichen and Huang, Xiande},
  journal={arXiv preprint arXiv:2505.12343},
  year={2025}
}

@inproceedings{jiang2025devils,
  title={Devils in middle layers of large vision-language models: Interpreting, detecting and mitigating object hallucinations via attention lens},
  author={Jiang, Zhangqi and Chen, Junkai and Zhu, Beier and Luo, Tingjin and Shen, Yankun and Yang, Xu},
  booktitle={Proceedings of the IEEE/CVF Conference on Computer Vision and Pattern Recognition},
  pages={25004--25014},
  year={2025}
}

@inproceedings{liu2025reducing,
  title={Reducing hallucinations in large vision-language models via latent space steering},
  author={Liu, Sheng and Ye, Haotian and Zou, James},
  booktitle={The Thirteenth International Conference on Learning Representations},
  year={2025}
}

@inproceedings{wu2025sharp,
  title={Sharp: Steering hallucination in lvlms via representation engineering},
  author={Wu, Junfei and Ding, Yue and Liu, Guofan and Xia, Tianze and Huang, Ziyue and Sui, Dianbo and Liu, Qiang and Wu, Shu and Wang, Liang and Tan, Tieniu},
  booktitle={Proceedings of the 2025 Conference on Empirical Methods in Natural Language Processing},
  pages={14357--14372},
  year={2025}
}

\end{document}